\documentclass[runningheads]{llncs}
\usepackage{graphicx}
\usepackage{amssymb} 
\usepackage{amsfonts} 
\usepackage{amsmath}

\usepackage{url}

\renewcommand{\vec}{\textbf}

\newcommand{\specificECE}{ECE^{c}}

\newcommand{\confidenceECE}{ECE^{conf}}

\newcommand{\classwiseECE}{ECE^{cw}}

\newcommand{\multiclassECE}{ECE^{mul}}

\newcommand{\probability}{\mathbb{P}}
\newcommand{\expectation}{\mathbb{E}}

\newcommand{\simplex}{\Delta}
\DeclareMathOperator*{\argmax}{argmax}

\newcommand{\model}{M}

\newcommand{\classes}{C}

\begin{document}
\title{Estimating Expected Calibration Errors}
%
%
\author{Nicolas Posocco\inst{1} \and
Antoine Bonnefoy\inst{1}}
\authorrunning{N. Posocco et al.}
%
\institute{EURA NOVA, Marseille, France\\
\email{firstname.lastname@euranova.eu}}
\maketitle              
\begin{abstract}
Uncertainty in probabilistic classifiers predictions is a key concern when models are used to support human decision making, in broader probabilistic pipelines or when sensitive automatic decisions have to be taken.
Studies have shown that most models are not intrinsically well calibrated, meaning that their decision scores are not consistent with posterior probabilities. 
Hence being able to calibrate these models, or enforce calibration while learning them, has regained interest in recent literature. 
In this context, properly assessing calibration is paramount to quantify new contributions tackling calibration.
However, there is room for improvement for commonly used metrics and evaluation of calibration could benefit from deeper analyses.
Thus this paper focuses on the empirical evaluation of calibration metrics in the context of classification.
More specifically it evaluates different estimators of the Expected Calibration Error ($ECE$), amongst which legacy estimators and some novel ones, proposed in this paper.
We build an empirical procedure to quantify the quality of these $ECE$ estimators, and use it to decide which estimator should be used in practice for different settings.

\end{abstract}

\section{Introduction}
\label{intro}

Almost all currently used classifiers are not intrinsically well-calibrated \cite{Niculescu-Mizil2005}, which means their output scores can't be interpreted as probabilities. This is an issue when the model is used for decision making, as a component in a more general probabilistic pipeline, or simply when one needs a quantification of the uncertainty in model's predictions, for example in high risk applications.

To overcome this calibration issue, two main tracks have been explored by either correcting the calibration of the model via some post-training procedure \cite{Platt1999,Niculescu-Mizil2005,Kull2017,Kull2019} or by regularizing the model to enforce calibration during training \cite{Kumar2018}. Would it be for the quantitative comparison of the performances of calibration methods or the evaluation of prediction's uncertainty, one needs to precisely quantify calibration. The recent literature trend is to use estimators of the Expected Calibration Error ($ECE$) \cite{Naeini2015}, which we focus on in this work.

We propose a few improvements on current $ECE$ estimators as well as a novel approach for the estimation of this metric based on kernel density estimation. We also introduce via these new estimators a continuous equivalent of the reliability diagram constructed on the proposed notion of Local Calibration Error ($LCE$). This notion can be used in practice to evaluate the uncertainty of the predicted probabilities itself, with an optional uncertainty interval. Furthermore we designed the first experimental setup to enable the assessment of the calibration metrics, in order to identify which estimators are the most relevant.

In this paper, we first present the context of this study in Section \ref{sec:related_works} and set up the formal definition of calibration in Section \ref{sec:calibration}. The theoretical calibration metric, namely the Expected Calibration Error, and its legacy and newly proposed estimators, are presented in Section \ref{sec:estimationofcalibrationquantification}, where we also introduce the concept of Local Calibration Error. We finally assess in Section \ref{sec:empiricalsetup} the relevance of legacy and proposed estimators empirically using a broad empirical setup.\footnote{The code ensuring the reproducibility of the experiments presented in this work is available at \url{https://github.com/euranova/estimating_eces}}

\section{Context and Related Work}
\label{sec:related_works}

The oldest attempt to quantify calibration has been the reliability diagram \cite{DeGroot1983,Niculescu-Mizil2005} for binary classification. Although it has been useful for the evaluation of early calibration methods, it does not provide point estimates - a single value - required to systematically compare calibration of different models. The first point estimate proposed in \cite{Zadrozny2001}, which exploited a decision theory framework to use a profit maximisation as a proxy for calibration quality, required a specific type of dataset to be usable in practice.
Mirroring the procedure used to compute the reliability diagram, the empirical Expected Calibration Error ($ECE$) was designed \cite{DeGroot1983}, and later has been proven to be an estimator for the natural theoretical notion of calibration error \cite{Guo2017}. Meanwhile, some works have used the negative log-likelihood (NLL) or the Brier score \cite{Zadrozny2001}, which both are weak proxis for the calibration of classifiers \cite{KullFlash2015}. Using reliability diagrams has become even more difficult in multiclass settings \cite{Zadrozny2002}.

Recent works mostly rely on the binning based legacy estimator of the $ECE$ to quantify calibration. Defects have been highlighted with this estimator, such as its reliance on a hyperparameter and its bias variance trade-off \cite{Nixon2019}. More recently \cite{Kull2019} made clearer the notion of calibration for multiclass classifiers, and new estimators of the $ECE$ with adaptive binning have been proposed in \cite{Nixon2019} along side with uncertainty aware reliability diagrams \cite{Brocker2007}. Although the notion of calibration was originally defined for classifiers, this notion is currently being generalized to regression \cite{Keren2018,Song2019}.

In this context we aim at improving the evaluation of calibration in the setting of classification, and specifically focus on estimators of the ECE as the theoretical definition itself has been consistently adopted. 

\begin{definition}{Classifier}

Let us consider the random variable $(X,Y)$, from which are drawn i.i.d samples to build a training set, and a holdout set of size $N$ : $(\vec{x},y) \in \mathcal{X} \times [1..C]$.
A classifier $M : \mathcal{X} \rightarrow \simplex^\classes$ is a function learnt from the training set which outputs scores -ideally the probabilities $\probability(Y | X)$- of belonging to class $c$ for $c\in[1..\classes]$, where $\simplex^\classes \triangleq \{\vec{s} \in [0,1]^\classes| \sum_{j=1}^\classes \vec{s}_j = 1\}$ is the probability simplex that ensures the scores sum up to one. 
In the rest of the paper the indexed notation $\vec{s}_c$ represents the $c$\textsuperscript{th} element of any vector $\vec{s}$ and $\vec{x}^i$ denotes the $i$\textsuperscript{th} sample of the holdout set. For readability purpose we use the notation $\vec{s}^i$ for the output score $M(\vec{x}^i)$.
\end{definition}

\section{Calibration}
\label{sec:calibration}



In this section we present and formalize properly the 4 different notions of calibration, and derive the corresponding Expected Calibration Errors ($ECE$). 

Calibration characterizes how much a model is able to output scores corresponding to actual posterior probabilities.
The first and simplest calibration notion \cite{Platt1999} is focused on a specific class and extends to the simultaneous calibration of every classes considering their associated scores independent, namely the class-wise calibration \cite{Zadrozny2002}. This version considers a classifier is well-calibrated if all one-vs-rest submodels are calibrated. The calibration concept for binary classification is equivalent to class-specific calibration focusing on the positive class and
to class-wise calibration, since the score for the negative class $s_0$ is determined by the score for the positive class $s_1 = 1 - s_0$. 
The more recently introduced \emph{confidence calibration} \cite{Guo2017} is only concerned about the model predicting relevant scores for the class it predicts for each sample. 
Throughout this paper, we only tackle the confidence and class-wise settings.
Finally the most rigorous evaluation of calibration should actually take into account all classes as non-independent, the corresponding definition, the \emph{multiclass-calibration} \cite{Platt1999} is almost never used in practice for computability reasons. All these notions are formalized in the following definition.

\begin{definition}{Different calibration notions of a probabilistic classifier.} A probabilistic classifier $M$, is
\begin{align*}
\text{Calibrated for class $c$: } &
\forall s \in [0, 1], \probability(Y = c | M(X)_c = s) = s  \\
\text{Class-wise calibrated: } & 
\forall c \in [1..C], \forall s \in [0, 1], \probability(Y = c | M(X)_c = s) = s \\
\text{Confidence-calibrated: } & 
{\forall s \in\! [0, 1], \probability(Y\!=\!\argmax_{c \in [1..C]}(M\!(X)_c) |\max_{c \in [1..C]}(M\!(X)_c)\! =\! s)\! =\! s} \\
\text{Multiclass-calibrated: } &
\forall \vec{s} \in \simplex^\classes, \forall c \in [1..\classes], \probability(Y = c | M(X) = \vec{s}) = \vec{s}_c 
\end{align*}
\end{definition}

The Expected Calibration Error ($ECE$) of a given model $\model$ can be naturally derived from these theoretical formulations by computing the expected deviation from the perfect theoretical calibration. 
This concept is applied to the different calibration settings and results in the following formulations:

\begin{definition}\label{def:ece}Expected calibration error ($ECE$) for the different settings for a given model M on $(X,Y)$:
\begin{eqnarray*}
    & \specificECE(\model) & \triangleq  \expectation_{s \sim \model(X)_c} \Big[\big|\probability(Y = c\ |\ \model(X)_c = s) - s\big| \Big]\\
    & \classwiseECE(\model) & \triangleq  \frac{1}{C}\sum_{c \in [1..C]} \specificECE(\model)\\
    & \confidenceECE(\model) & \triangleq  \expectation_{s \sim \max(\model(X))} \Big[\big|\probability(Y = \argmax(\model(X))\ |\ \max(\model(X)) = s) - s\big| \Big]\\
    & \multiclassECE(\model) & \triangleq  \expectation_{(\vec{s}, c) \sim ({\model(X)}, Y)} \Big[\big|\probability(Y = c | \model(X) = \vec{s}) - \vec{s}_c \big| \Big]
\end{eqnarray*}


Where $\specificECE(\model)$ is the class-specific $ECE$ associated to class $c$, $\classwiseECE$ the class-wise $ECE$ \cite{Zadrozny2002}, $\confidenceECE(\model)$ the confidence $ECE$ \cite{Guo2017} and $\multiclassECE(\model)$ the multiclass $ECE$.

\end{definition}






By replacing the expectation over the absolute values of the differences by a simple maximum over the absolute differences, we obtain the formulations of the Maximum Calibration Error ($MCE$) \cite{Naeini2015}, which focus on the highest gap between posterior probabilities and the scores given by the model.

\section{Estimation of calibration quantification}
\label{sec:estimationofcalibrationquantification}

In this section we describe the challenges of calibration quantification, then present the existing tools to handle these challenges namely the reliability diagram and the legacy $ECE$ estimator. 
We then introduce a new formalization of these estimators based on binning and sample mapping, which help us define new binning based estimators. Finally we present the new notion of Local Calibration Error on which we rely to build continuous estimators of the ECE based on Kernel Density Estimation. All estimators are written for the class-specific calibration setting, which can then be transposed to the other settings using Definition \ref{def:ece}.

\subsection{Challenges of such quantification}
\label{sec:estimationofcalibrationquantification_challengesofsuchquantification}

Quantifying calibration is challenging in practice for two main reasons:
    \emph{Calibration is intrinsically a local notion.} Miscalibration is defined on the neighbourhood of a given output score. Thus any global quantification of calibration depends on an aggregation procedure of local measures. This is what differentiates the $ECE$, which implicitly weights all parts of the score distribution according to its local density, from the $MCE$, which only cares about the worst case scenario.
    \emph{Since calibration depends on score distributions }, any relevant estimator relies on these scores, which means that we are limited by the amount of available validation data to perform such quantification.

A good calibration metric should \emph{specifically quantify calibration}: contrary to the Brier score and the NLL, which values only carry a partial information on calibration, we expect a good metric to be independent of confusion factors. It should then be \emph{theoretically well-funded} as well as \emph{tractable in practice}. Finally, a good calibration metric should be able to \emph{take into account cost matrices for the classification task}, when available, risk management being intrinsically linked to such cost matrices.

The $ECE$ corresponds to the identified required properties for homogeneous cost matrices, since it directly derives from the theoretical notion of calibration and has an immediate interpretation. However, it doesn't allow heterogeneous costs matrices, and as we will see in the next sections, current estimators provide poor estimations of the true value of the $ECE$.
For these reasons we focus on the setting of homogeneous cost classification, and try to provide better estimators for the $ECE$.
Such estimators should \emph{be robust to hyperparameter choice}, problem which can be solved by the use of a relevant heuristic. The estimator should be \emph{data-efficient} too, in order to provide good estimates with a \emph{low variance} even with few holdout labeled data points. Such estimation should provide \emph{low-bias} estimates with a sufficient amount of available data and should finally be \emph{consistent} and  \emph{computable in a reasonable amount of time}.

\subsection{Reliability Diagram}
\label{sec:estimationofcalibrationquantification_estimatorsoftheece}

The reliability diagram introduces the classical way of calculating the $ECE$.
To build the reliability diagram (in the binary setting), a uniform binning scheme (the $[0,1]$ interval is split into equal bins) is used, and each holdout sample is mapped into a bin based on the score given by the model for the positive class (procedure defined below as 1-bin mapping). For each bin, the average score for the positive class and the proportion of samples belonging to the positive class are calculated. The first is then plotted against the second. If the model is well calibrated, each point should fall on the line $y = x$. The local offset of each point tells us if the model is locally over or under-confident on its scores for the positive class. Such diagram can be seen on Figure \ref{fig:reliability_diagrams}~(left).

Originally designed for the binary classification case, it can be easily extended to confidence calibration in the multiclass setting. In that case, samples are sent into bins based on the score the model outputs for the class it predicts, and the ratio of correct predictions is plotted against the average over the scores given for the predicted class.

\subsection{Binning based estimators}

In order to present different binning-based estimators of the $ECE$, we formalize the binning and affectation mapping objects. We note $s^i$ the score of the class of interest of the $i$\textsuperscript{th} sample, which depends on if we consider the specific-class, class-wise (fixed class) or confidence (predicted class) calibration.

\begin{definition}{Binning schemes}\\
\label{def:binning_schemes}
The $[0,1]$ segment is split into $B$ bins used to assign each data point to one (or more) bin. These bins are defined by their respective thresholds. Hence to define a binning scheme one only needs to specify the increasing splitting function $t : [1..B] \rightarrow [0, 1]$ that computes the right threshold for each bin.
\end{definition}

Two main binning schemes have been used to compute the $ECE$ in the literature:
\emph{Uniform binning} splits the segment into $B$ bins of equal size : $t(i) = \frac{i}{B}$ and
\emph{Adaptive binning} splits the segment so that each split contains the same number of samples : $t(i) = \{s^{\sigma(j)}\ |\ \forall j \in [1.. N-1], s^{\sigma(j)} \leq s^{\sigma(j + 1)} \}_{\lfloor N/i \rfloor}$, $\sigma$ being the permutation which sorts samples based on the score predicted for the class of interest.

\begin{definition}{Affectation mapping}\\
\label{def:affectation_schemes}
Given a binning of a domain $[0, 1]$, an affectation mapping of $\mathcal{D}$ in these bins is a matrix $W$ composed of positive weights, so that $W_{ij}$ is the weight of the affectation of the sample $i$ in the bin $b_j$. Rows of such matrix sum up to $1$.
\end{definition}

Using this formalisation, we start from the \emph{1-bin mapping} $W^{1bin}$ for which every sample is assigned to a single bin with unit weight, to go to the new proposed \emph{convex mapping} $W^{conv}$ for which each sample may contribute to up to two bins for the computation of the binning based ECE estimators. This mechanism is the one referred to as \emph{linear binning} in the kernel density estimation field.
These two mappings can be respectively mathematically written, as follows, where $c_j$ is the geometric centre of the $j$\textsuperscript{th} bin $\frac{t_{j-1} - t_j}{2}$:

\noindent\begin{minipage}{.40\linewidth}
\begin{equation*}
\label{eq:1bin}
  W_{ij}^{1bin}= 
\begin{cases}
    1  &  \text{if } s_i \in [ t_{j-1}, t_j ]\\
    0  &  \text{otherwise}
\end{cases};
\end{equation*}
\end{minipage}%
\begin{minipage}{.53\linewidth}
\begin{equation*}
\label{eq:convexbin}
  W_{ij}^{conv}= 
\begin{cases}
    1 & \text{if } s_i \in [ 0, c_0 ] \text{ \& } j=0\\
    1 & \text{if } s_i \in [ c_{B}, 1 ] \text{ \& } j=B\\
    \frac{s_i - c_j }{c_{j+1} - c_j} & \text{if } s_i \in [ c_j, c_{j+1} ]\\
    1 - \frac{s_i - c_{j-1} }{c_{j} - c_{j-1}} & \text{if } s_i \in [ c_{j-1}, c_j ]
\end{cases}
\end{equation*}
\end{minipage}

The original estimator of the $ECE$ is basically a weighted mean over the absolute differences calculated when the reliability diagram is computed (here expressed in the specific-class case). If $W$ is a 1-bin mapping on a uniform binning and $\emph{1}$ is the indicator function, the legacy estimator is:

\begin{align}
\label{eq:legacyECE}
ECE_l^c = \frac{1}{N} \sum_{j=1}^{B} \Big| \sum_{i = 1}^N W_{ij} (\emph{1}_{Y^i = c} -  \vec{s}_c^i) \Big|
\end{align}
	
 Such estimator can be defined in the same way for $\confidenceECE$ and $\classwiseECE$. 
 
We unify binning-based estimators under equation (\ref{eq:legacyECE}) with different binning/mapping schemes. The $ECE_a$ uses an adaptive binning with 1-bin mapping, while the $ECE_c$ uses a uniform binning and a convex mapping, and finally the $ECE_{ac}$ uses both improvements on the legacy estimator - adaptive binning and convex mapping. In the case of class-wise calibration, the $ACE$ defined in \cite{Nixon2019} is equivalent to $ECE_a$, when all bins contain the same amount of samples.

\subsection{Local Calibration Error}
\label{sec:LCE}

We define the notion of Local Calibration Error ($LCE$), and then use it to build the reliability curve, a continuous version of the reliability diagram. 
Let us first begin with the formal definition of the LCE:

\begin{definition}{Local calibration error ($LCE$) for the class-specific and the confidence settings for a given model M on $(X,Y)$}
\begin{eqnarray*}
    LCE^{c}_\model(s) & \triangleq & \probability(Y = c\ |\ \model(X)_c = s) - s\\
    LCE^{conf}_\model(s) & \triangleq & \probability(Y = \argmax(\model(X))\ |\ \max(\model(X)) = s) - s
\end{eqnarray*}
\end{definition}

For the class-specific case, to estimate the $LCE$ of a model for all scores $s \in [0, 1]$, we have to estimate $\probability(Y = c|\model(X)_c = s)$. We resort to the Bayes rule to tear down this estimation to estimating the densities of $\probability(\model(X)_c = s|Y = c)$ and $\probability(\model(X)_c = s)$, and the scalar $\probability(Y=c)$. We can then rely on kernel density estimation (KDE) to estimate the two densities. Theoretically, this approach is continuous. In our implementation however, both KDEs are evaluated numerically in Fourier space (the first one on all scores for the class $c$ and the second one on all scores for the class $c$ when the ground truth is the class $c$), which makes the computation efficient with $O(N + n log(n))$ complexity, if $n$ is the number of numeric subdivisions of the domain $[0,1]$. We use steps of 0.0003 for precision, and mirrored the data around $s = 0$ and $s = 1$, which are the limits of the domain. This mirroring implies a slight bias in estimations due to a leak of density mass. Once again the $LCE^{conf}$ can be estimated in the same way with the relevant scores and classes.

\begin{figure}[t]
    \centering
    \includegraphics[width=\textwidth]{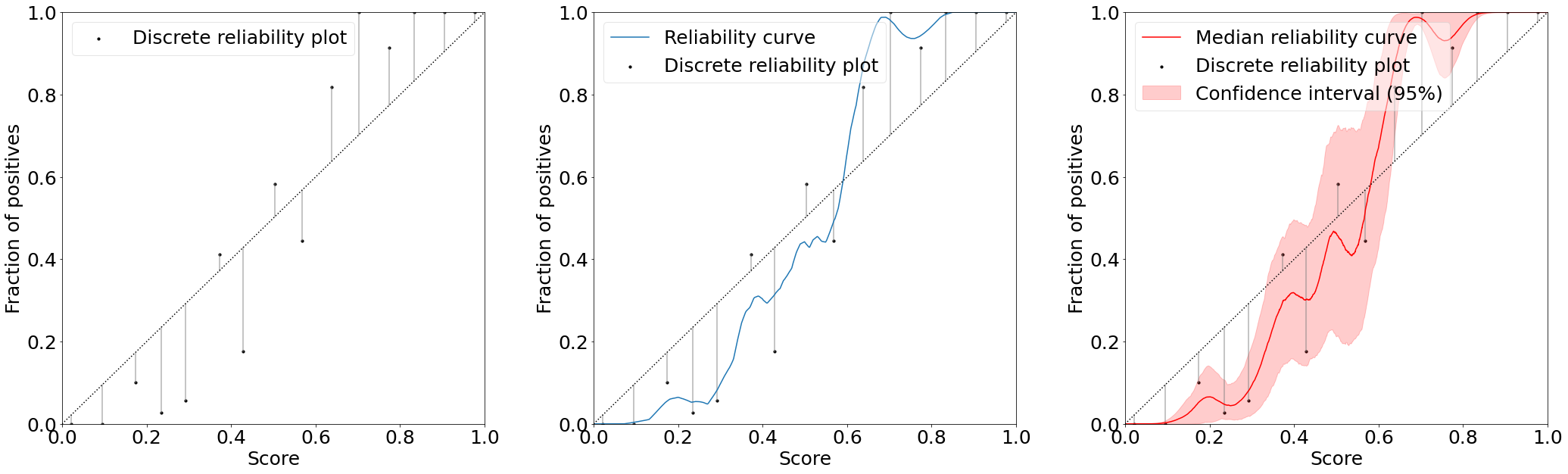}
    \caption{Reliability diagram with 15 bins (left), reliability curve with a bandwidth of 0.03 (middle) and the bootstrapped version with the same bandwidth (right). Each plot brings one more level of insight.}
    \label{fig:reliability_diagrams}
\end{figure}

A continuous equivalent of the reliability diagram can be derived from such object. The reliability curve associated with the classifier $\model$ and the class $c$, for the class-specific calibration is:
\begin{align}
\forall\ s_c \in [0,1], rel_\model(s_c) = LCE^c_\model(s_c) + s_c
\end{align}
An example of such reliability curve is shown in Figure \ref{fig:reliability_diagrams}~(middle).

The main benefit this proposed notion of local calibration error offers is its usability in practice to know the uncertainty of a model on a specific score, which cannot be evaluated with enough precision using previous tools (points in a reliability diagram can be used for an interpolation aiming at the same result, yet the precision of such procedure is very low, and interpolation at that scale is questionable). 



We propose to compute this curve on bootstrapped versions of the holdout set, in order to quantify the uncertainty on this $LCE$. In this context, the median curve is considered as the reliability curve and percentiles of interest are used for uncertainty quantification. This idea, illustrated on Figure \ref{fig:reliability_diagrams}~(right), allows the prediction of confidence intervals for the class probabilities instead of point estimates, by only looking at the uncertainty on the bootstrapped reliability curve at the score output by the model.



\subsection{Density based estimator: $ECE_d$}
\label{sec:estimationofcalibrationquantification_estimatorsoftheece_ece_kernel}

Based on the definition of this Local Calibration Error we can derive a new $ECE$ estimator, which is formalized as follows:

$$ECE_{d}^c = \int_0^1 f_{\model(X)_c}(s)\ | LCE_\model(s) |\ ds, $$
where  $f_{\model(X)_c}$ is the probability density function of the scores given by the model for class $c$.

\subsubsection{Heuristics for hyperparameter choices}
\label{sec:estimationofcalibrationquantification_estimatorsoftheece_heuristicsforhyperparameterchoices}

For all binning-based estimators, we investigate the use of a simple heuristic to select the number of bins used for the estimations: the bin amount is the square root of the number of samples. For the kde-based approach, we propose to use Silverman's rule \cite{silverman1986density} to select the bandwidth (the bandwidth is estimated on $\probability(\model(X)_c = s)$, and the same bandwidth is used to estimate the density of $\probability(\model(X)_c = s|Y = c)$). Other heuristics are often used for KDE computations, yet Silverman's rule is to our knowledge the only one which provides satisfying results in small data contexts, for which legacy estimators struggle the most.

\subsubsection{From class-specific to the other settings}
\label{sec:estimationofcalibrationquantification_estimatorsoftheece_fromclassspecifictoothersettings}

To translate the class-specific estimators into the class-wise case, class-specific ECEs are estimated for all classes, and the class-wise ECE is the mean of these values. To get to the confidence case, scores for the class of interest are replaced by the score for the predicted class, and the class of interest is the ground truth label.

\section{Experimental setup}
\label{sec:empiricalsetup}

We present the assessment of a few empirical properties of the different $ECE$ estimators. As pointed out in \cite{Nixon2019}, the main difficulty with empirical evaluation of calibration methods and calibration metrics is that we don't have access to ground truths in general. 
This is why we worked on a setup which gives us access to arbitrarily precise estimates of the $ECE$ considered as a the ground truth, in the class-wise and confidence settings.

\subsection{Procedure}

We aim at quantitatively compare the estimators in terms of approximation, data efficiency and variance. To do so we build curves which can indicate the expected performance of each estimator with its corresponding parameters, for different sizes of holdout set. 
In order to observe statistically robust result we introduce various degrees of variability in our experiment at distribution level, in the algorithm used to train the models, and in terms of train/holdout sets splits. The results are thus produced based on numerous realistic output score distributions.

The distribution variability is introduced by creating synthetic sample sets from Gaussian mixtures, where each class is composed of 4 modes of the mixture. For each mode we build the mean vector with elements uniformly drawn in $[0,1]$, and the covariance matrix is built as follows: we first sample a matrix with elements uniformly drawn in $[-0.3, 0.3]$ then multiply it with its transposition to get the required positive definite matrix. This sample set generation is produced with various number of classes (${2, 5, 7}$) and dimensions of the feature space (${2, 5, 7}$) with 5 different large datasets sampled from each combination, resulting into 45 synthetic distributions.

\begin{figure}[t]
    \centering
    \includegraphics[width=\textwidth]{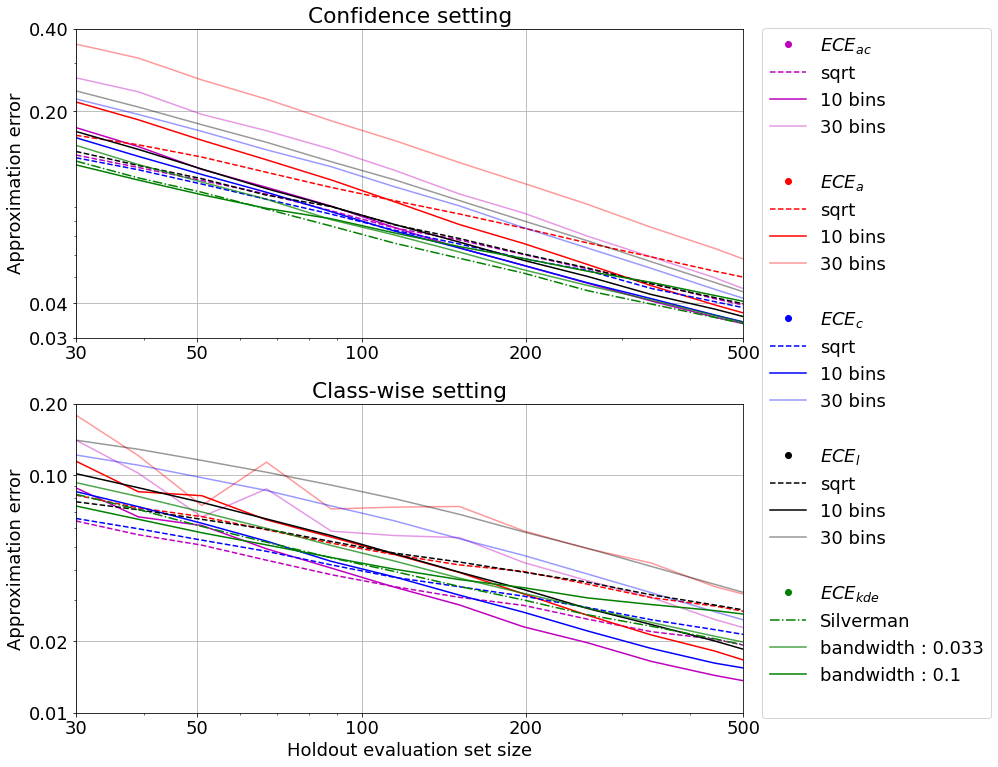}
    \caption{Median 95th percentile of the approximation error (absolute value of the normalized relative deviation with respect to the ground truth $ECE$) for the different estimators of the $\confidenceECE$ (top) and the $\classwiseECE$ (bottom) (lower is better) for evaluation sets of size between 30 and 500 samples. The scale is logarithmic for both axis}
    \label{fig:setup}
\end{figure}

In order to produce various relevant score distributions from these data distributions, we trained 4 different types of models (logistic regression, gaussian naïve bayes classifier, support vector classifier and random forest) on 3 train sets of size 300 sampled out from the previously generated large datasets. 
For each of these trained models we compute the "ground truth" $ECE$ using the legacy estimator with high granularity (2000 bins) on the remaining holdout set ($2.10^6$ samples). 
Then, we build 200 evaluation sets which are bootstrapped versions of the holdout set of sizes taken between 30 and 500 on a logarithmic scale. 
The "ground truth" $ECE$ is used as reference to compute the approximation error (the absolute value of the difference between the estimated $ECE$ and its true value normalized by the ground truth). Among those 200 values per evaluation set size, we keep the 95$^{th}$ percentile of the approximation errors, below which 95\% of such errors rely. 
For each evaluation set size and estimator, we finally plot the median over the 540 95\textsuperscript{th} percentiles obtained with each score distributions. The resulting curves can be seen in Figure \ref{fig:setup}. 
The number of evaluations of the learning algorithms plus the ECE estimators makes this experiment long to run, but as all the estimators have limited computation complexity the overall computation remains feasible.

\subsection{Results analysis}

\paragraph{For all settings,} the error of all estimators is very high for small data regimes (the estimation error is around 10\% of the true value), and thus one shouldn't evaluate calibration on so little data, no matter what estimator is used.


\paragraph{For the confidence setting,} the best performing estimator in almost all data regimes is the $ECE^{conf}_{d}$ with Silverman's rule. This is good news, since we now have a procedure to estimate the ECE which doesn't rely on a sensitive hyperparameter choice, but instead on a simple heuristic. As far as it is concerned, the convex mapping scheme empirically improves the performance of the legacy estimator and the one using the adaptive binning, which underperforms when alone, probably because of the increased variance induced by the adaptivity. It is worth noting that above 300 samples, a lot of the estimators show similar performances. As far as the square root heuristic is concerned for the automatic choice of number of bins used, the graphs suggest that the number of bins grows slightly too fast in average with an increasing amount of samples.
\paragraph{For the class-wise setting,} there is no clear outperformer in all data regimes among the tested estimators. For less than 100 samples, the $ECE_{ac}^{cw}$ with the square root heuristic seems to be the best choice. The same estimator, this time with a fixed small number of bins, is then the most precise one. The observation made earlier about the square root heuristic still holds, and Silverman's heuristic for the bandwidth seems to be a less relevant choice in the class-wise setting than in the confidence one. We assume it is the case because of the sharpness of the score distributions for each classes in the class-wise setting (most of the density being very close to 0 and 1), which is a context in which Silverman's bandwidth is known to underperform for kernel density estimation.

\section{Conclusions}

We have introduced a few improvements on the legacy estimators, from the proposition of new binning schemes to the use of heuristics to automatically pick relevant values for hyperparameters of estimators of the ECE. On top of this, a novel approaches has been built to define properly the notion of local calibration error, which produces novel estimators for the ECEs. By testing all approaches on a synthetic experimental setup for which we had access to very precise estimates of the theoretical ECE, we have been able to compare all candidate estimators. This systematic evaluation, which had never been done until now, allowed us to formulate some recommendations on which estimator to use in what context.

Our proposed solutions lead to natural potential future works. First, the introduced calibration curve suggests a natural post-training calibration method, since it can be seen as a calibration map. Such method would be interesting to evaluate, yet poses the problem that the associated calibration maps are not monotonous, which is considered as a prerequisite for post-hoc calibration procedures in the literature. Then multiclass-calibration evaluation, which is still an open problem today, could potentially be evaluated in the scores space using an adapted variant of our kde approach, which we think wouldn't suffer as much as legacy estimators from the increase of dimensionality. Finally, even if this paper uses classical kernels and a mirroring approach to constrain density estimations on the domain $[0,1]$ which allows standard and fast KDE computation, some preliminary investigations using a beta pseudo-kernel (the second one introduced in \cite{Chen1999}) which is naturally constrained to this domain, show promising results. Because this kernel has a different shape for all support points in $[0,1]$, it is computationally prohibitive for now, and needs further exploration.

\bibliographystyle{splncs04}
\bibliography{biblio}

\end{document}